# Continental-scale habitat distribution modelling with multimodal earth observation foundation models

**Authors**

Sara Si-Moussi[1*], Stephan Hennekens[2], Sander Mucher[2], Stan Los[2], Yoann Cartier[3,4], Borja Jiménez-Alfaro[5], Fabio Attorre[6], Jens-Christian Svenning[7] and Wilfried Thuiller[1]

1. Univ. Grenoble Alpes, Univ. Savoie Mont Blanc, CNRS, LECA, F-38000 Grenoble.
2. Wageningen Environmental Research (WENR), part of Wageningen University and Research (WUR), PO Box 47, 6700 AA, Wageningen, The Netherlands.
3. LEESU, ENPC, Institut Polytechnique de Paris, Univ Paris Est Créteil, Marne-la-Vallée, France
4. LHSV, ENPC, Institut Polytechnique de Paris, EDF R&D, Chatou, France
5. IMIB Biodiversity Research Institute (Univ. Oviedo-CSIC-Princ. Asturias), University of Oviedo, Oviedo, Spain
6. Sapienza University of Rome, Department of Environmental Biology, P.le Aldo Moro 5, 75 00185 Rome, Italy
7. Center for Ecological Dynamics in a Novel Biosphere (ECONOVO), Department of Biology, Aarhus University, Ny Munkegade 114, DK-8000 Aarhus C, Denmark

*** Corresponding author:**
Sara Si-Moussi sara.si-moussi@univ-grenoble-alpes.fr

Address:
UFR Chimie & Biologie – Laboratoire d'Ecologie Alpine - Bâtiment D, 2233 Rue de la Piscine, 38610 Gières

## Abstract

Habitats integrate the abiotic conditions, vegetation composition and structure that support biodiversity and sustain nature's contributions to people. Most habitats face mounting pressures from human activities, which requires accurate, high-resolution habitat mapping for effective conservation and restoration. Yet, current habitat maps often fall short in thematic or spatial resolution because they must (1) model several mutually exclusive habitat types that co-occur across landscapes and (2) cope with severe class imbalance that complicates exhaustive multi-class training.

Here, we evaluated how high-resolution remote sensing (RS) data and Artificial Intelligence (AI) tools can improve habitat mapping across large geographical extents at fine spatial and thematic resolution. Using vegetation plots from the European Vegetation Archive, we modelled the distribution of Level 3 EUNIS habitat types across Europe and assessed multiple modelling strategies against independent validation datasets.

Strategies that exploited the hierarchical nature of habitat classifications resolved classification ambiguities, especially in fragmented habitats. Integrating satellite-borne multi-spectral and radar imagery, particularly through Earth Observation (EO) Foundation models (EO-FMs), enhanced within-formation discrimination and overall performance. Finally, ensemble machine learning that corrects class imbalance boosted predictive accuracy even further.

Our methodological framework is transferable beyond Europe and adaptable to other classification systems. Future research should advance temporal modelling of habitat dynamics, extend to habitat segmentation and quality assessment, and exploit next-generation EO data paired with higher-quality in-situ observations.

**Keywords**

Multimodal EO, geospatial AI, foundation models, ensemble ML, Habitat distribution models, EUNIS, ecosystem extent

# 1. Introduction

Habitats are fundamental to biodiversity: they host taxonomically, phylogenetically, and functionally rich biota and sustain complex ecological interactions (Gaüzère et al., 2022; Pollock et al., 2020) that underpin a wide range of economic, cultural, and societal services (Díaz et al., 2015). Yet, they are under growing threat from climate change, pollution, land-use intensification, and invasive species, leading to widespread degradation and loss (AEE., 2019; Brondízio et al., 2019; Rogers et al., 2023).

In the context of monitoring and conservation, habitats are not understood in the abstract sense of the proximal conditions in which species live or their ecological niches, but rather as discrete ecological entities jointly defined by abiotic factors (e.g., climate, topography) and the characteristic vegetation communities they support (Davies et al., 2005; Mucina et al., 2016). This definition underpins the EUNIS habitat classification in Europe (Chytrý et al., 2020) and is broadly equivalent to ecosystem types in the IUCN Global Ecosystem Typology (Keith et al., 2022), providing a common framework for comparing and monitoring habitats across regions.

Mapping these habitat types is central to biodiversity monitoring because it informs on ecosystem extent, an Essential Biodiversity Variable (EBV) with direct relevance for policy and management (Geijzendorffer et al., 2016; Jetz et al., 2019; Pereira et al., 2013). Effective habitat maps must meet three criteria: (1) high thematic resolution, to move beyond broad land-cover classes (e.g., forest, grassland, shrubland) and capture ecologically meaningful units (Evans, 2010; Keith et al., 2015); (2) high spatial resolution, to detect habitat patches and assess fragmentation (Wintle et al., 2019); and (3) broad spatial extent, to enable cross-border conservation and coordinated management (European Commission. Joint Research Centre., 2020; Hermoso et al., 2022).

The pressing demand for such habitat maps has spurred advances in habitat distribution modelling (EEA & MNHN, 2014). Habitat distribution models (HDM) are statistical approaches that predict the likelihood of a habitat type based on environmental predictors, treating habitat mapping as a classification problem. To model multiple habitat types, researchers have investigated different classification schemes in the literature.

Many studies adopted a binary classification scheme, by building individual habitat suitability models (HSMs) for each habitat class (C. A. Mücher et al., 2009; S. Mücher & Hennekens, 2018), often leveraging species distribution modelling (SDM) techniques (Guisan et al., 2017; Rapinel & Hubert-Moy, 2021) and software such as MaxEnt (Phillips et al., 2017) and biomod2 (Thuiller et al., 2009, 2016). By modelling habitat classes independently, these approaches ignore the spatial exclusivity and co-occurrence of habitats leading to inflated estimates of the spatial extent covered by each habitat.

Multi-class classification approaches simultaneously model all habitat classes to predict the most probable type. This scheme has been assessed at limited spatial or thematic resolutions (Agrillo et al., 2021, 2021; Giannetti et al., 2018; Huber et al., 2023; Immitzer et al., 2019; Lafitte et al., 2024; Le Dez et al., 2021; Marzialetti et al., 2019) owing to the complexity of modelling habitat distribution over large geographic areas.

Some habitats are mutually exclusive due to the distinct ecological requirements of their plant communities (e.g., mediterranean vs alpine heathlands), others coexist within the same spatial unit (e.g., riparian forests with river waterbodies, orchards with scattered trees and pasture), and many exhibit short-term temporal dynamics such as seasonal changes (e.g., temporary waterbodies alternating between aquatic and terrestrial states) or land use driven rotations (e.g., pasture–cropland cycles). Such spatial and temporal dynamics create ambiguity in multi-class habitat classification.

Remote sensing (RS) provides a time series of spatially contiguous images that is needed to capture habitat dynamics over large spatial extents (Ali et al., 2020; O'Connor et al., 2015; Vihervaara et al., 2017). Beyond broad land-cover mapping (Malinowski et al., 2020; Venter et al., 2022; Zanaga et al., 2022), RS data can supply biologically meaningful features such as vegetation indices and phenological metrics, which characterize fine-scale habitat properties (Lausch et al., 2016; Skidmore et al., 2021; Ustin & Middleton, 2021). Recently, Earth Observation (EO) foundation models (EO-FM) (Szwarcman et al., 2025; Y. Wang et al., 2025; Xiao et al., 2024) that are deep neural networks pretrained on vast satellite archives using self-supervised learning techniques, have been shown to capture spatio-temporal dynamics of satellite imagery across diverse ecosystems potentially unveiling fine-grained ecological structure. While successful in tasks such as land-use classification, the application of EO-FMs to fine-grained habitat distribution models remains unexplored.

Despite these advances, no study has systematically assessed RS-driven modelling approaches for large-scale, high-resolution habitat classification across all habitat types. We propose that artificial intelligence (AI) techniques offer new opportunities: deep learning can extract spatial patterns directly from imagery, while machine learning captures complex interactions among abiotic and biophysical predictors.

Yet major challenges remain: (1) few studies systematically compare classification schemes on independent datasets (Burns et al., 2022); (2) the hierarchical structure of habitat typologies such as EUNIS are underutilized (Gavish et al., 2018); and (3) habitat types distribution is strongly imbalanced: some habitats (e.g., lowland hay meadows) are widespread, whereas others (e.g., Canarian xerophytic scrub) may have restricted ranges due to specific environmental constraints or historical land use change (Álvarez-Martínez et al., 2018; Mucina et al., 2016). This class imbalance poses statistical challenges (Benkendorf et al., 2023; Krawczyk, 2016; Sidumo et al., 2022) leading to a mapping focus on common habitats.

This study presents the first large-scale comparative assessment of AI-powered approaches relying on RS data for habitat distribution modelling. Our objectives are twofold:

1. To evaluate the contribution of different RS products to habitat mapping at fine thematic and high spatial resolutions.
2. To provide methodological guidelines to optimize AI tools for large-scale habitat distribution modelling.

To address these questions, we modelled EUNIS Level 3 habitats across Europe using the European Vegetation Archive (EVA) database. We systematically compared different modelling methodologies on independent validation datasets, offering insights into the feasibility and accuracy of AI-based habitat classification at continental scale (Figure 2).

# 2. Methods

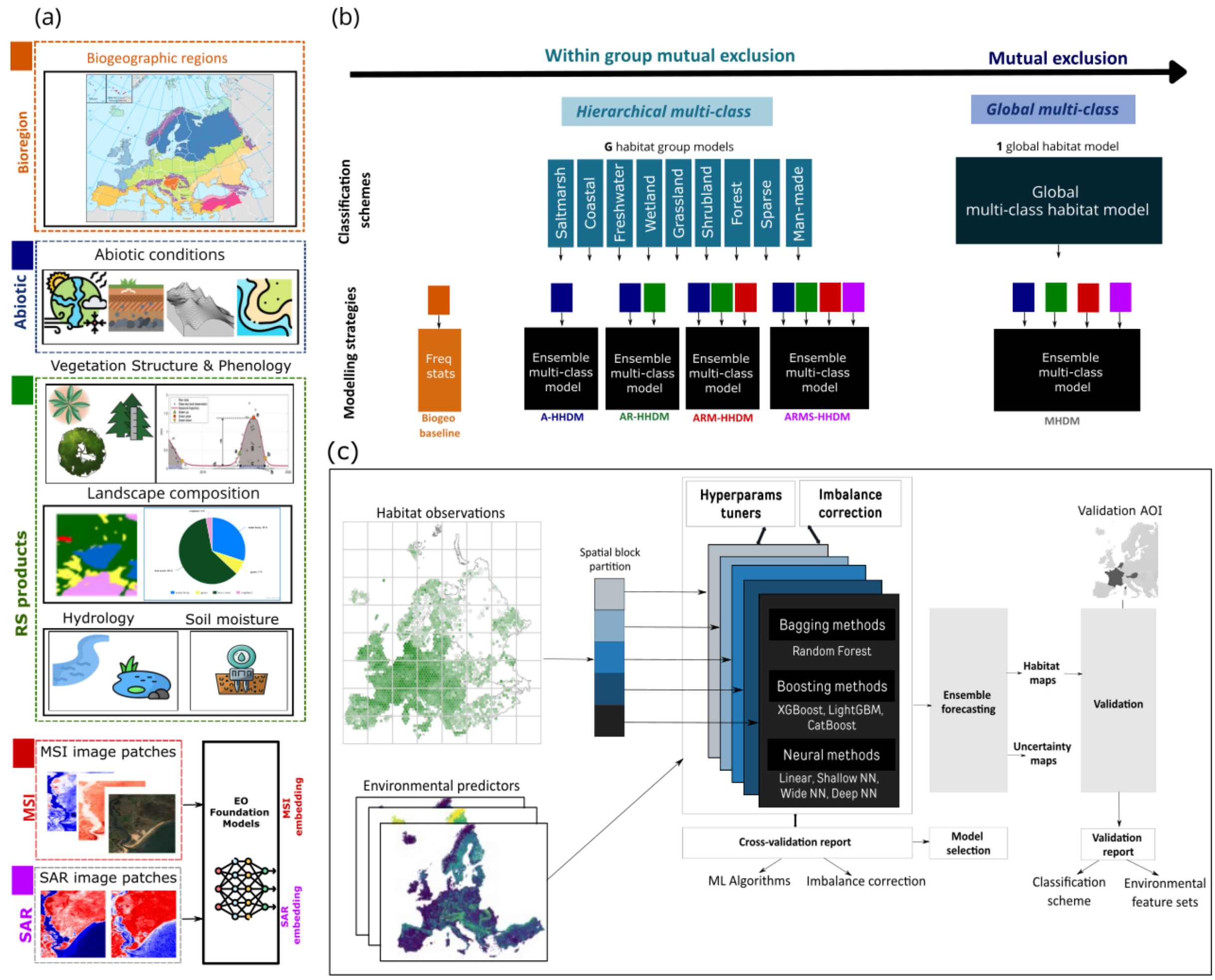

*Figure 1 - Overview of the habitat distribution modelling strategies compared in this study.*

*Panel (a) presents the environmental modalities used as model inputs including biogeographic regions (orange), abiotic conditions such as climate, soil and terrain (blue), remote sensing (RS) products providing both biotic and abiotic features such as vegetation structure, phenology, landscape composition, hydrology, and soil moisture (green), multispectral imagery (MSI; red) and synthetic aperture radar (SAR; purple) processed into respective embeddings using Earth Observation (EO) foundation models (EO-FMs).*

*Panel (b) illustrates the conceptual design of the multiple habitat distribution modelling strategies we conducted in this study, structured along a gradient of co-occurrence constraints, ranging from a hierarchical multi-class model within habitat groups (partial mutual exclusion) to a global multi-class model representing full mutual exclusivity. Each classification scheme is combined with different environmental modalities, denoted by color-coded squares that correspond to panel (a) yielding various modelling strategies. Ensemble models are used for each strategy.*

*Panel (c) details the ensemble machine learning framework, which integrates habitat observations through a spatial block partitioning of the entire study area for cross-validation, and diverse environmental predictors. Various machine learning algorithms including decision tree ensembles such as bagging (e.g., Random Forest) and boosting algorithms (e.g., XGBoost, LightGBM, CatBoost), multi-layer perceptron neural networks (e.g., shallow, wide, and deep architectures), and tabular transformers, are tuned with hyperparameter optimization and imbalance correction methods. The framework outputs ensemble-based habitat class probabilities and associated uncertainty maps. Habitat distribution predictions are validated against independent habitat distribution observations.*

## 2.1. Habitat observations

European habitat distributions were modelled at 100-m resolution using the EUNIS Habitat Classification. We worked at level 3 of the classification hierarchy, which distinguishes 249 habitat types (Chytrý et al., 2020) nested within nine broad formations defined at level 1 including saltmarshes (MA2), coastal habitats (N), freshwater vegetation (P), wetlands (Q), grasslands (R), shrublands (S), forests (T), sparsely vegetated habitats (U) and man-made vegetation habitats (V).

### 2.1.1. Training data

To train habitat distribution models, we used vegetation plot data from the European Vegetation Archive (EVA) (Chytrý et al., 2016; European Vegetation Survey, The IAVS Working Group, 2024). Only vegetation plots collected after 2000, with a plot area between 1m² and 1000m², georeferenced with a positional uncertainty below 100m were retained. Vegetation plots were classified into EUNIS habitat types at level 3 using the EUNIS-ESy expert system (Chytrý et al., 2021). Habitat classes with less than 5 observations after duplicates removal were discarded. The resulting dataset contained a total of 597,810 georeferenced habitat observations covering 243 habitat classes.

Table 1 summarizes for each habitat formation (EUNIS level 1), the number of habitat classes at level 3 and the total number of plots. Additionally, we report the imbalance ratio in the training dataset computed as the prevalence ratio of the most frequent to the least frequent level 3 classes within the level 1 formation.

*Table 1 - Number of habitat classes, vegetation plots and imbalance ratio (ratio of frequency of most prevalent to rarest habitat class in the training data) for each EUNIS level 1 habitat formation.*

| Habitat formation (level 1) | Saltmarshes (MA2) | Coastal (N) | Freshwater (P) | Wetlands (Q) | Grassland (R) | Shrubland (S) | Forest (T) | Sparsely vegetated (U) | Man-made (V) |
|---|---|---|---|---|---|---|---|---|---|
| N° classes | 11 | 25 | 10 | 20 | 54 | 43 | 46 | 26 | 12 |
| N° plots | 6552 | 20287 | 10452 | 57764 | 217522 | 39341 | 186842 | 5889 | 53161 |
| Imbalance ratio | 59.9 | 205.6 | 66.9 | 702.6 | 4273.8 | 340.6 | 2180.0 | 148.9 | 584.2 |

To avoid over-optimistic estimates of model predictive performances that arise when spatially close samples are split between training and test sets, all observations were assigned to spatial blocks in a 100x100km grid (see section 1.4). We used the official EEA grid. These blocks were then partitioned into 5 folds for cross-validation. This procedure ensures that evaluation always took place in areas unused during model training (Figure 1c). As an illustration, Figure 2 illustrates the partitions obtained considering all habitat classes for all formations.

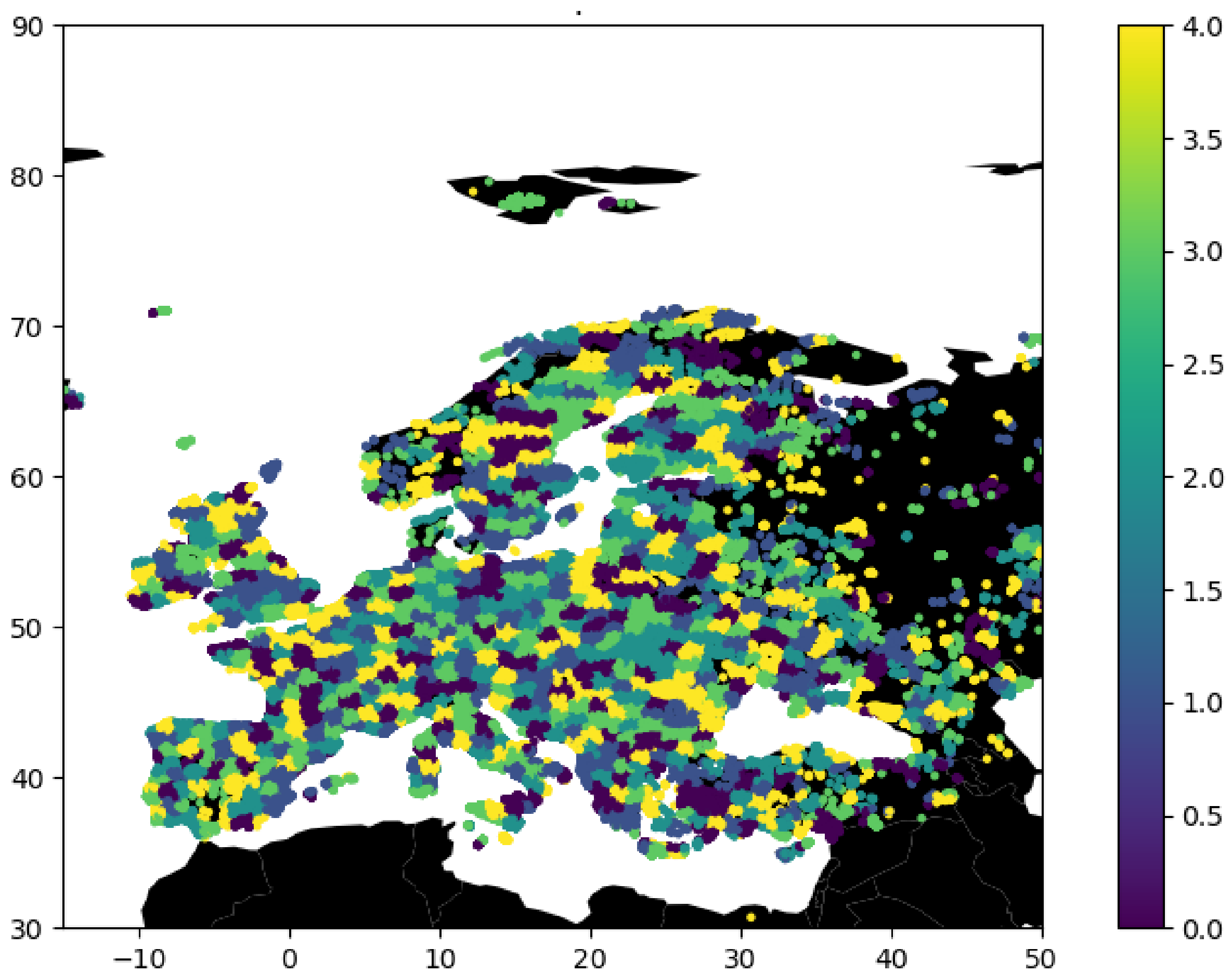

*Figure 2 - Spatial distribution of training plots and their partitioning into the 5 spatial block cross-validation folds (color) for the global multi-class habitat distribution model strategy (MHDM).*

### 2.1.2. Validation data

To assess the accuracy of the predicted habitat maps, we relied on two independent sources of habitat occurrence data and two regional habitat maps.

The first dataset (NL) comprises 49,512 vegetation plots collected between 2010 and 2022 across the Netherlands from the Landelijke Vegetatie Database (LVD) (Hennekens, 2018; Schaminée et al., 2012) and classified into EUNIS habitats following a similar procedure to EVA plots.

The second dataset (PT) comprises 150,003 habitat polygons from Southern Portugal generated by applying expert-driven crosswalk from Portugal’s 2007 land-cover map (COS) and vegetation series to EUNIS habitats for southern Portugal (Mesquita et al., 2021).

The third dataset (IFN) includes 21,252 forest plots sampled between 2013 and 2021 in Eastern France (Grand-Est, Vosges, Jura, southern Alps) as part of the French Forest Inventory (IFN – Inventaire Forestier National Français, 2022), a long-term national monitoring program that provides annual data on forest conditions. For each plot, the habitat at the central point is determined with regional identification keys and harmonized with EUNIS at level 3. Finally, we used the MAES/EUNIS habitat map for Austria (AT) (Umweltbundesamt, 2021). This product integrates biotope mapping datasets from federal states, harmonized to EUNIS Level 3, into a nationwide 10-m resolution raster (≈ 8 million classified cells).

## 2.2. Environmental data

We defined five sets of predictors (modalities) to characterize environmental conditions relevant to habitat distribution, depicted in Figure 1a.

### 2.2.1. Environmental modalities

- The **Bioregion set:** Each vegetation plot was labelled with one of the 13 European biogeographic regions recognised by the European Environment Agency. This categorical variable captures long-term evolutionary and historical constraints on regional species pools and therefore on habitat occurrence. The map of biogeographic regions of Europe at 100m resolution was used as the reference grid.
- The **Abiotic (ABIO)** set: It combines climate, edaphic and topographic descriptors (Table 2) compiled at the highest continent-wide resolution available (100 m to 1 km). Climate variables were derived from the CHELSA database (Karger et al., 2017, 2020) including annual mean temperature, temperature seasonality, annual precipitation, precipitation seasonality, snow cover and growing-degree days averaged over a climatic period (1990-2020). Topography variables were derived from the Copernicus Digital Elevation Model (European Space Agency & Airbus, 2022) including aspect, ruggedness, land-form classes and distance to the coastline. Geological context was described with the geological class from the Global Lithological Map (Hartmann & Moosdorf, 2012) and the depth-to-bedrock from the European Soil Database (Panagos et al., 2022). Finally, we extracted physical and chemical soil indicators from SoilGrids (Poggio et al., 2021), including texture fractions, bulk density, pH, organic carbon, nitrogen and cation-exchange capacity. All abiotic predictor layers were resampled at 100m and aligned to the reference grid.
- The **Remote Sensing Products (RSP)** set: To complement the abiotic set, we incorporated remote-sensing products describing present-day ecosystem structure, composition and function (Table 3). Vegetation structure was represented by canopy height and density, summer and spring leaf area index, and fractional land-cover proportions (Zanaga et al., 2022). The hydrographic context combined river-network density (European Environment Agency, 2019), surface-water inundation frequency (Pekel et al., 2016) and binary masks of permanent water, wetlands and ice/snow (Buchhorn et al., 2020). Functional dynamics were summarized through PPI-based phenological metrics (season length, amplitude, green-up and senescence slopes, maximum productivity) (Copernicus Land Monitoring Service, 2021) and two indices of surface-soil moisture (Copernicus Land Monitoring Service & Copernicus Land Monitoring Service, 2018). All the layers were resampled and aligned with our reference grid. Remote sensing predictors spanned different periods depending on availability of EO products. To reduce interannual variability and short-term noise, we averaged them over their respective ranges to capture stable environmental regimes.

- The **Optical (multi-spectral) image patches (MSI)** set: For every vegetation plot, we extracted a 256 × 256-pixel Sentinel-2 patch (≈ 2 km side length) centred on the plot (Figure 3). Annual, cloud-free median mosaics for year 2020 were used and all nine bands available at 10 m (B02-B08, B11, B12) were retained. This mosaic includes: (i) visible bands Blue (B2), Green (B3) and Red (B4) for basic landcover discrimination; (ii) Red edge bands (B5, B6, B7) that are sensitive to chlorophyll content and canopy structure aiding in detection of vegetation condition and plant functional types; (iii) Near Infrared (NIR) (B8) for biomass and vegetation health; and (iv) Short-wave Infrared (SWIR) (B11, B12) for water content and bare soil (Clevers & Gitelson, 2013; Delegido et al., 2011; Drusch et al., 2012). The spatial context contained in these patches allows the model to exploit neighbourhood configuration beyond the immediate pixel.
- The **Synthetic Aperture Radar (SAR)** set: An analogous 256 × 256-pixel patch was built from Sentinel-1 dual-polarisation (VV, VH) backscatter composites for 2020, radiometrically terrain-corrected and expressed as $\gamma^0$ (Figure 3). Processed via GAMMA software, the composite includes VV and VH polarizations median-aggregated over a full year. VV polarization detects surface roughness, distinguishing built-up and open areas, while VH cross-polarization captures vegetation volume scattering which is prevalent in vegetation canopies due to the interaction of the radar signal with leaves, branches, and stems. SAR complements optical data by capturing structural (surface roughness and vegetation volume) and dielectric (moisture content) land surface properties independently of weather and lighting conditions, ensuring reliable monitoring in persistently cloudy regions (Torres et al., 2012).

*Table 2 - List of abiotic predictors (ABIO) used for habitat modeling.*

| Group | Predictor | Data source |
|---|---|---|
| Climate | **BIO1**: Annual mean temperature (°C) | CHELSA (Karger et al., 2017, 2020)<br>Original resolution: 1km<br>Temporal range: 1990-2020<br>URL |
| | **BIO4** : Temperature seasonality – bio 4 | |
| | **GDD5**: Growing degree days heat sum above 5°C (°C) | |
| | **BIO12**: Annual sum of precipitation (kg.m$^2$/yr) | |
| | **BIO15**: Precipitation seasonality | |
| | **SCD**: Snow covered days (day count) | |
| | **SWE**: Snow water equivalent (kg.m$^2$/yr) | |
| Terrain | **TRI**: Topographic ruggedness index | Copernicus Digital Elevation Model (DEM) (European Space Agency & Airbus, 2022)<br>Original resolution: 30m<br>Temporal range: 2011-2015<br>URL |
| | **ASPECT**: EU DEM aspect (degrees) | |
| | **TPI_LANDFORM**: Landform classification based on the topographic position index, using SAGA-GIS *ta_morphometry* toolbox (10 classes) | |
| | **COAST**: whether location is in a coastal area (5km buffer around European coastlines) | European Environmental Agency (EEA)<br>Original resolution: 100m<br>Temporal range: 2017<br>URL |
| Geology | **GLIM_CLASS**: Dominant parent material class (16 classes) | Global Lithological Map (GLiM) (Hartmann & Moosdorf, 2012)<br>Original resolution: 1km<br>Temporal range: N/A<br>URL |
| | **DR:** Depth to rock (m) | European Soil Database (ESDB) (Panagos et al., 2022)<br>Original resolution: 1km<br>Temporal range: 2006<br>URL |
| Soil | **BDOD**: Bulk density of the topsoil (kg/dm$^3$) | Soilgrids (Poggio et al., 2021)<br>Original resolution: 250m<br>Temporal range: 1960-2020<br>URL |
| | **Sand**, **Silt**, and **Clay** proportions (%) | |
| | **pH**: Soil acidity | |
| | **SOC**: Organic carbon content (g/kg) | |
| | **N**: Nitrogen content (g/kg) | |
| | **CEC**: Cation exchange capacity (cmol/kg) | |

*Table 3 - Remote sensing biodiversity products (RSP) used for habitat modeling.*

| Group | Predictor | Data source |
|---|---|---|
| Vegetation phenology and productivity | **AMPL:** Season amplitude given by MAXV-MINV | CLMS (Copernicus Land Monitoring Service, 2021)<br>Original resolution: 10m<br>Temporal range: 2016-2021<br>URL |
| | **LENGTH:** Length of season (number of days between start and end) | |
| | **LSLOPE:** Slope of the greening phase (growth) of the growing season (*PPI × day*-1) | |
| | **RSLOPE:** Slope of the browning phase (senescence) of the growing season (*PPI × day*-1) | |
| | **MAXV**: PPI at the day of maximum-of-season | |
| | **TPROD**: Total productivity (PPI × day) | |
| Vegetation structure | Leaf area index in summer ($m^2/m^2$) | CLMS (Copernicus Land Monitoring Service, 2017)<br>Original resolution: 300m<br>Temporal range: 2014-2021<br>URL |
| | Leaf area index in spring ($m^2/m^2$) | |
| | Tree canopy cover density (%) | CLMS (Copernicus Land Monitoring Service, 2025)<br>Original resolution: 100m<br>Temporal range: 2012-2021<br>URL |
| | Height of the tree canopy (m) | ETH Zurich (Lang et al., 2022, 2023)<br>Original resolution: 10m<br>Temporal range: 2020<br>URL |
| Hydrography | **INUNDATION_OCC**: Number of inundation days over a year | JRC, Global Surface Water (Pekel et al., 2016)<br>Original resolution: 30m<br>Temporal range : 1984-2021<br>URL |
| | **WATER_BODY:** binary presence of water body on a 100m radius | Copernicus Global Land Service (Buchhorn et al., 2020)<br>Original resolution: 100m<br>Temporal range: 2015-2019<br>URL |
| | **WETLAND:** binary presence of a wetland on a 100m radius | |
| | **ICE_GLACIER_SNOW:** binary presence of glacier on a 100m radius | |
| | **RIVER_DENSITY:** density of water courses weighted by their hydrological importance | EU-Hydro<br>Original resolution: 100m<br>Temporal range: 2006-2012<br>URL |
| Soil moisture | **SWI_TOPSOIL**: Topsoil wetness index | Copernicus land monitoring service (Copernicus Land Monitoring Service & Copernicus Land Monitoring Service Helpdesk, 2018)<br>Original resolution: 1km<br>Temporal range: 2015-2021<br>URL |
| | **SOIL_MOISTURE** : Soil surface moisture | Original resolution: 1km<br>Temporal range: 2014-2021<br>URL |
| Fractional land cover | Proportion (%) of pixels of each landcover class in a 100m radius (except water bodies and wetlands) | ESA WorldCover (Zanaga et al., 2022)<br>Original resolution: 10m<br>Temporal coverage: 2020-2021<br>URL |

For the ABIO and RSP variables, a single value was extracted at the precise location corresponding to each habitat observation. In contrast, MSI and SAR image patches were utilized to capture a broader spatial context surrounding each observation (Figure 3). This methodology facilitated the analysis of the spatial configuration of landscape elements, thereby complementing conventional metrics of landscape composition. Our predictors dataset therefore combines tabular (ABIO, RSP) and image (MSI, SAR) modalities.

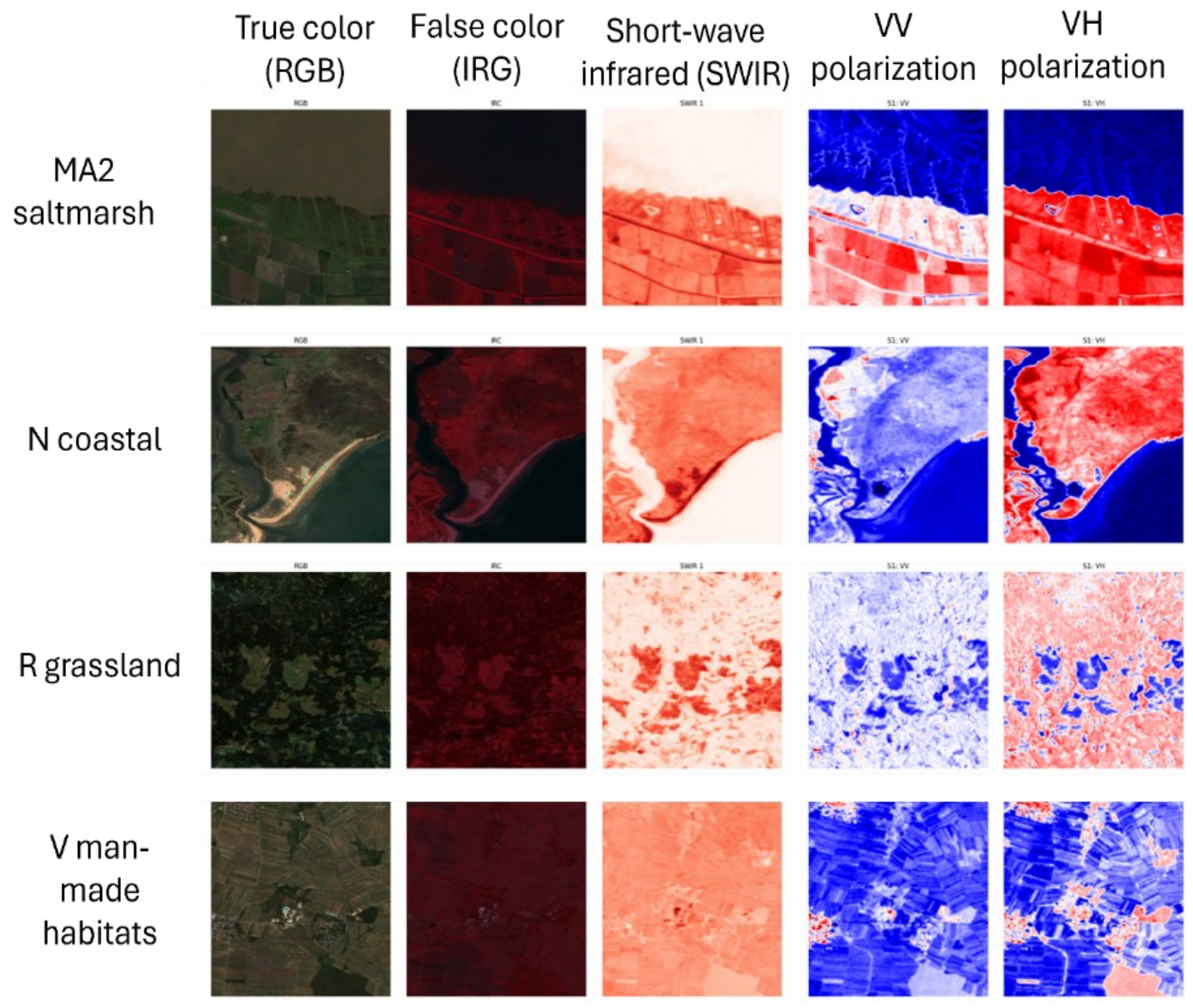

*Figure 3 - Example of MSI and SAR image patches for different habitat types*

## 2.2.2. MSI and SAR imagery representations

### 2.2.2.1. EO-FMs

We used EO-FMs to extract one-dimensional feature vectors (embeddings) from MSI and SAR image patches. We evaluated a diverse set of EO-FMs for multi-spectral image embedding based on various neural architectures including convolutional neural networks (CNN) and vision transformers (ViT) and pretrained with diverse self-supervised learning strategies including contrastive learning, self-distillation and masked image modelling, on global datasets (see Table 4). Finally, for radar imagery, we only evaluated the Dynamic One-For-All (DOFA) foundation model.

*Table 4 - List of Earth Observation Foundation Models evaluated in this study*

| Architecture | EOF Model | Pretraining Strategy | Spatial Scope | Supported Sensors |
|---|---|---|---|---|
| **CNN** (ResNet-50) | SSL4EO (Y. Wang et al., 2023) | MoCo / Dino | Europe | Sentinel-2 |
| | SSL4ECO (Plekhanova et al., 2025) | Seasonal Contrast | Global | Sentinel-2 |
| | EO4B (our baseline) | Supervised (habitat classification) | Europe | Sentinel-2 |
| **ViT** (ViT-Base) | SSL4EO (Y. Wang et al., 2023) | MoCo / Dino | Europe | Sentinel-2 |
| | DOFA (Xiong et al., 2024) | Masked Image Modeling | Global | Sentinel-1, Sentinel-2, NAIP, EnMAP |
| | Prithvi (Szwarcman et al., 2025) | Masked Image Modeling | Europe | Sentinel-2 |

#### 2.2.2.2. Supervised baseline (EO4B)

We assessed whether embeddings obtained from self-supervised learning are more relevant than those obtained from a supervised model (EO4B) trained on a related task of predicting EUNIS habitat formations given MSI imagery.

##### Data

Multiple EUNIS level 1 formations can co-occur within a patch. The task was thus formulated as a multi-label classification problem. Multi-label annotations were generated by complementing habitat labels at level 1 with ancillary spatial layers describing fractional land cover based on high resolution layers (Worldcover, (Zanaga et al., 2022), dominant land cover (Corine land cover, CLC) as well as delimitations of coastal and water areas (European Environment Agency, 2019). Specifically, coastal habitats were assigned within coastal areas, freshwater near water bodies, saltmarshes where the CLC class “saltmarshes” was present, and wetlands where the CLC class “inland wetlands” occurred. Terrestrial formations (grasslands, scrub, forests) were complemented when vegetation cover exceeded 10 %, and unvegetated or man-made classes when bare or built surfaces surpassed 10%.

#### Model architecture

EO4B builds on a ResNet-50 V2 backbone pretrained on the BigEarthNet dataset, providing domain-specific initialization for Sentinel-2 imagery. The final embedding layer (2048 features) was followed by a fully connected layer of 1024 ReLU units with batch normalization and a sigmoid output layer of nine nodes.

#### Training

Image patches were normalized using global statistics and augmented by random horizontal and vertical flips ($p = 0.5$). The model was trained end-to-end using focal binary cross-entropy ($\gamma = 5.0$) to mitigate label imbalance, optimized with AdamW (weight decay = 0.001) and a cosine-annealed learning-rate scheduler. Training followed an 80/10/10 stratified split (train/validation/test) with early stopping to prevent overfitting. The architecture was implemented in PyTorch Lightning and trained for 100 epochs on 4 × NVIDIA V100 GPUs.

*Table 5 – Test performances of the CNN-based EUNIS level 1 multi-label classifier (EO4B)*

| Label | ROC_AUC | Sensitivity | Specificity | TSS |
|---|---|---|---|---|
| MA2 | 0.97 | 0.92 | 0.93 | 0.85 |
| N | 0.98 | 0.93 | 0.92 | 0.85 |
| P | 0.86 | 0.76 | 0.80 | 0.56 |
| Q | 0.89 | 0.85 | 0.77 | 0.62 |
| R | 0.81 | 0.75 | 0.71 | 0.46 |
| S | 0.75 | 0.63 | 0.73 | 0.37 |
| T | 0.75 | 0.64 | 0.71 | 0.36 |
| U | 0.90 | 0.80 | 0.84 | 0.64 |
| V | 0.72 | 0.68 | 0.63 | 0.32 |

EO4B achieved high discrimination of EUNIS level 1 formations (ROC-AUC = 0.72–0.98; TSS = 0.32–0.85) on the test set (Table 5) and provides a transferable CNN encoder generating landscape embeddings for downstream ecological prediction and habitat-mapping tasks.

### 2.2.2.3. Dimensionality reduction

For each EO-FM and for EO4B, we applied Principal Component Analysis (PCA) to reduce the dimensionality of the embeddings up to 100 components and ~90% cumulative explained variance. The resulting PCs were concatenated with tabular predictors and used as input for the learning algorithms.

## 2.3. Modelling framework

### 2.3.1. Classification schemes

Because multiple habitats may co-occur in the same spatial unit (here 100 m by 100 m), we compared two multi-class paradigms:

- In the global multi-class habitat distribution model (MHDM), all 249 classes are considered simultaneously and are forced to be mutually exclusive at every pixel.
- In the hierarchical multi-class habitat distribution model (HHDM), exclusivity is applied within each of the nine broad formations; classes from different formations (such as alpine grassland and coniferous forest) can be predicted simultaneously. This approach generates nine formation-specific classifiers.

### 2.3.2. Multi-modal modelling strategies

First, the global approach (MHDM) used all modalities (ABIO + RSP + MSI + SAR). This approach assumes that remote sensing-based descriptors (RSP, MSI, and SAR) can differentiate among habitats that co-occur under similar abiotic conditions, independently of their broad habitat formation.

To limit the computational cost of the study, all technical and sensitivity tests were carried under this global MHDM. As such, we used it to compare the different EO-FMs on MSI imagery, to test the sensitivity of the image size for extracting the spatial context from MSI and SAR, to test the overall contribution of the different environmental modalities through an ablation study, and finally to test various ways of accounting for class-imbalance. Once these technical tests were carried out, we used the best combinations of hyper-parameters, spatial context and class-imbalance selected, we carried out the comparison between the global model (MHDM) and its hierarchical counterpart (HHDM).

For the hierarchical approach (HHDM), and for the best combination of EO-FMs, class imbalance and spatial extent, we built four nested variants to quantify the added value of progressively richer RS information:

(i) *A-HHDM* with only ABIO;

(ii) *AR-HHDM* adding RSBIO;

(iii) *ARM-HHDM* further including MSI embeddings;

(iv) *ARMS-HHDM* finally adding SAR embeddings.

The latter two configurations test whether the inclusion of spatial context improves classification performance.

A naïve baseline, **Biogeo-HDM**, simply assigns to every plot the most frequent EUNIS level 3 class observed within the same formation and biogeographic region. It is used as a lower bound to discard low-performing models.

## 2.4. Learning algorithms and ensemble model

Each strategy was fitted inside a common machine learning framework that first screens four algorithmic families:
(1) Decision tree ensemble algorithms including bagging with RandomForest (Breiman, 2001) and boosting with XGBoost, CatBoost and LightGBM (Chen & Guestrin, 2016; Ke et al., 2017; Prokhorenkova et al., 2018).
(2) Multi-layer perceptron neural networks (MLP) setting one to three hidden layers (LeCun et al., 2015);
(3) Tabular transformers, with the tabular attentive network (TabNet) (Arik & Pfister, 2021) setting depth from one to five layers and width ranging between 8 and 64.

Finally, we combined their best instances in an ensemble modelling approach.

### 2.4.1. Imbalance correction

Given the imbalance in habitat distributions, we focused on techniques that do not involve reduction (e.g., under-sampling majority classes) or artificial inflation (e.g., over-sampling minority classes) of the data or synthetic data generation (e.g., SMOTE) to avoid exacerbating the biases in the data. Instead, we selected approaches that adapt the optimization objective to balance classes during training. We tested several approaches and techniques on the global MHDM.
For tree-based models (random forests and gradient boosting), we addressed class imbalance using class-weighting, scaling error penalties according to inverse class frequency. In neural networks, we adapted the loss function beyond the default categorical cross-entropy (CE). Specifically, we tested weighted CE (WCE) which increases the contribution of minority classes, Focal Loss (FL) which down-weights easy cases to focus on hard or misclassified samples (Lin et al., 2017), and the Label Distribution-Aware Margin (LDAM) loss which enlarges prediction margins for underrepresented classes (Cao et al., 2019). We also evaluated the combination of class weighting with FL and LDAM (wFL, wLDAM) to further reduce misclassification of rare habitats.

### 2.4.2. Training and model selection

All models were trained using 5-fold spatial block cross-validation (Roberts et al., 2017) to prevent data leakage between training and evaluation sets induced by spatial autocorrelation and ensure generalization across heterogeneous landscapes. We used the Iterative Stratification algorithm from *scikit-multi-learn* (Szymański & Kajdanowicz, 2018) to preserve class distribution across folds. Within each cross-validation fold, we implemented automated hyperparameter tuning procedures tailored to each algorithm on a 10% random hold-out of the training dataset. We adapted feature pre-processing routines to the feature types and seamlessly integrated them at the fold level to the training/prediction pipelines. Finally, we used the cross-validation performances to select the best algorithms within each family (bagging, boosting, neural networks, and tabular transformers) and their optimal imbalance correction techniques to compose the final ensemble model for each modelling strategy.

### 2.4.3. Ensemble forecasting

The ensemble model aggregates predictions from multiple algorithms and cross-validation folds using a weighted voting strategy, where models contribute based on their validation performance. This approach integrates algorithmic diversity and reduces variance, yielding robust predictions.

## 2.5. Statistical analyses

### 2.5.1. Evaluation metrics

Because the spatial blocking strategy may leave some classes unseen during training, our primary cross-validation metric is top-k accuracy, i.e., the proportion of pixels for which the true class appears among the *k* classes predicted with highest probability. We also report the coverage error: the average position of the true habitat class in the ranked list.

To evaluate the ensemble models on the validation datasets, we use class-wise recall, precision, and F1 scores. Recall (the proportion of habitat classes correctly identified) reflects omission errors, while precision (the proportion of predicted habitat classes that are correct) reflects commission errors. The F1 score is their harmonic mean, balancing precision and recall.

### 2.5.2. Comparison of EO-FMs to extract meaningful representations

To compare the information provided by the different EO-FMs and EO4B on the MHDMs, we performed the Friedman test on class-wise F1-scores obtained on the validation datasets. Following significant Friedman test results, we applied a post-hoc Nemenyi test to identify which pairs of strategies differed significantly. The Nemenyi test compares average ranks and controls for multiple comparisons. The best EO-FM based on F1-score was then used in subsequent analyses.

### 2.5.3. Spatial context sensitivity

While the comparison of the different EO-FMs effects on habitat predictions was carried out using an initial patch size of 128x128 pixel (~1 km), we then determined the optimal neighbourhood extent when accounting for the spatial context. For this, MSI/SAR encoders from the selected EO-FM were also recomputed with 64 × 64 and 256 × 256 patches (≈ 500 m and 2 km). The MHDM performance differences for the three spatial contexts (64 × 64, 128 x 128, and 256 × 256) were assessed with paired t-tests on F1 score, while all other parameters were held constant to isolate the effect of spatial context. The best combination of EO-FM and spatial context was then used for the final MHDM and for the HHDM comparisons.

### 2.5.4. Multi-modal ablation study

We performed an ablation study to evaluate the contribution of each environmental modality in our final global MHDM. Models were trained with one modality removed at a time, while maintaining identical hyperparameters and optimization settings. The impact of each modality was measured by comparing the coverage error of each variant to that of the full model.

### 2.5.5. Evaluation and comparison of MHDM and HHDM

For comparisons between classification schemes and for the four nested variants of HHDM, we evaluated class-wise recall, precision, and F1 scores from the ensemble predictions on the validation datasets. For the global model, these metrics were also computed within each broad formation to enable direct comparison with its hierarchical counterpart (HHDM).

### 2.5.6. Feature attribution

For the best model between MHDM and HHDM and for each habitat formation (EUNIS level 1), we computed relative feature contributions across the EVA dataset using the framework of Shapley Additive explanations (SHAP) (Lundberg & Lee, 2017) to identify the most important predictors for fine-grained habitat classification.

# 3. Results

## 3.1. Comparison of EO-FMs on the global multi-class strategy

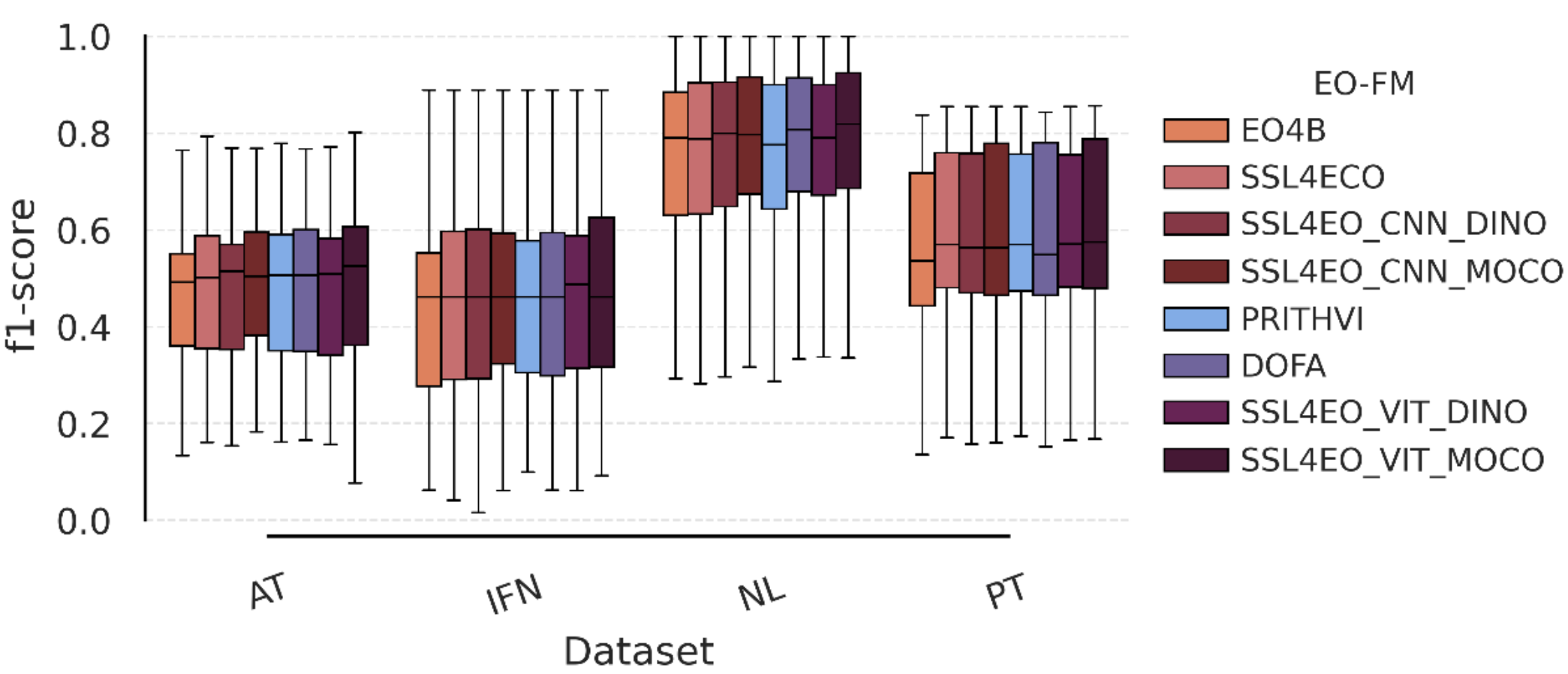

*Figure 4 Comparison of class-wise f1-score in the validation datasets across EO foundation models within the global multi-class strategy (MHDM).*

All self-supervised models (DOFA, Prithvi, SSL4EO variants, SECO) outperformed the supervised baseline (EO4B). Amongst EO-FMs, performance differences were very small and class-dependent (Figure 4) yet statistically significant across the validation datasets (Friedman test on f1-score, $\chi^2 = 346.98$, $p < 0.001$). The ranking of EO-FMs did neither depend on the self-supervised pretraining strategy nor on the encoder architecture, but rather on the chosen metric. For recall, Prithvi had the highest mean score, with DOFA, SSL4Eco, and several SSL4EO variants performing comparably. For precision, SSL4EO_VIT_MOCO led, while DOFA and SSL4EO_CNN_MOCO were not significantly different. DOFA was retained for downstream analyses given its competitive recall–precision balance, native support for multimodal (Sentinel-1/2) inputs and active maintenance (Y. Wang et al., 2025).

## 3.2. Sensitivity to spatial context size

When comparing the effects of the spatial context (≈ 500 m, 1 km and 2 km) around the plots covered by MSI and SAR images, we showed that while average differences in recall, precision, and F1-score were small, Friedman tests revealed significant variation across habitat classes (recall $\chi^2 = 12.04$, precision $\chi^2 = 22.08$, F1-score $\chi^2 = 21.15$). This was specifically the case for littoral (MA2), coastal (N), wetlands (Q) and man-made habitats (V), whereas differences were inconsistent for the other formations (Table 6). Post-hoc tests showed that a spatial context of 1 km (image size 128x128) yielded the best recall ($p = 0.029$ vs. 2 km), while 500 m (image size 64x64) performed best in precision ($p = 0.001$ vs. 2 km), indicating fewer false positives. In the following analyses, we thus kept the 128 × 128 patches (1 km).

*Table 4 - Summary of class-wise F1-score across habitat modelling strategies for each EUNIS level 1 formation with different image sizes. The optimal image size is marked with an asterisk *. All statistically equivalent to it are highlighted in bold.*

| Image Size | MA2 | N | P | Q | R | S | T | U | V |
|---|---|---|---|---|---|---|---|---|---|
| ~500m 64x64 | **0.967 ± 0.029*** | **0.867 ± 0.099*** | 0.904 ± 0.055 | **0.749 ± 0.075** | 0.688 ± 0.178 | 0.849 ± 0.086 | 0.746 ± 0.109 | 0.917 ± 0.083 | **0.760 ± 0.053*** |
| ~1km 128x128 | **0.962 ± 0.035** | **0.867 ± 0.095** | 0.901 ± 0.054 | **0.750 ± 0.076*** | 0.688 ± 0.175 | 0.846 ± 0.087 | 0.747 ± 0.109 | 0.916 ± 0.084 | **0.754 ± 0.049** |
| ~2km 256x256 | **0.958 ± 0.038** | 0.865 ± 0.095 | 0.895 ± 0.054 | 0.744 ± 0.077 | 0.687 ± 0.174 | 0.846 ± 0.089 | 0.743 ± 0.111 | 0.915 ± 0.083 | 0.750 ± 0.051 |

## 3.3. Multi-modal ablation effects

Ablation of specific modalities increased coverage error in all cases relative to the full and final MHDM, with the largest rise observed when removing the ABIO (+11.1%) and RSP sets (+7.2%). Multispectral imagery removal caused a moderate increase (+3.6%), while the ablation of the set of SAR predictors had minimal effect (+0.4%). In other words, the habitat classifiers maintained overall high performances despite missing modalities.

## 3.4. Model selection: algorithms and imbalance correction

Among individual algorithms, XGBoost, a three-layer multilayer perceptron, and TabNet performed similarly, achieving top-3 accuracies close to 0.81–0.82 and top-5 accuracies around 0.88. The performance-weighted ensemble of the four algorithm families exceeded all single learners, reaching 0.83 (top-3), 0.89 (top-5) and the lowest coverage error (3.18 ± 0.30; Table 7).

*Table 5 - Spatial block cross-validation predictive performances (top 3 and top 5 accuracy, coverage error) for the best performing ML algorithms and their ensemble mean.*

| Algorithm | TOP3 accuracy | TOP5 accuracy | Coverage error |
|---|---|---|---|
| **XGBoost** | 0.81 ± 0.03 | 0.88 ± 0.02 | 3.34 ± 0.31 |
| **MLP** | 0.81 ± 0.03 | 0.88 ± 0.02 | 4.41 ± 0.52 |
| **TabNet** | 0.80 ± 0.03 | 0.87 ± 0.02 | 4.51 ± 0.91 |
| **Ensemble** | **0.83 ± 0.03** | **0.89 ± 0.02** | **3.18 ± 0.30** |

For XGBoost, simple class weighting improved the macro-averaged F1, but when the imbalance was extreme, as in the grassland formation (R), this strategy resulted in lower recall for common classes decreasing the overall score. In neural networks, replacing the standard cross-entropy with focal loss or LDAM yielded clear benefits, and in several formations an additional class-weighting term further improved recall without penalising precision (notably MA2, N, P and Q).

## 3.5. Comparative performance of habitat-modelling strategies

Table 8 summarizes the class-wise validation F1 score obtained broken down by the nine EUNIS level-1 formations across modelling strategies. We provide the class-wise scores across strategies in the Supplementary Information. All strategies out-performed the purely biogeographic baseline. The hierarchical model that used the full predictor stack (ARMS-HHDM) delivered the highest F1 in six formations and never ranked below second place on the rest. Particularly large gains over the global model were observed in wetlands (Q) and grasslands (R), which contain many visually similar classes that benefit from the nested formulation of exclusivity.

*Table 6 - Summary of class-wise F1-score across habitat modelling strategies for each EUNIS level 1 formation across the validation datasets. The best strategy is marked with an asterisk *. All strategies statistically equivalent to it are highlighted in bold.*

| | Habitat formations (EUNIS level 1) | | | | | | | | |
|---|---|---|---|---|---|---|---|---|---|
| Strategy | MA2 Salt marshes | N Coastal | P Freshwater | Q Wetland | R Grassland | S Shrub-land | T Forest | U Sparse vegetation | V Man-made |
| BIOGEO | 0.366 ± 0.423 | 0.087 ± 0.192 | 0.101 ± 0.214 | 0.059 ± 0.152 | 0.049 ± 0.181 | 0.077 ± 0.202 | 0.072 ± 0.190 | 0.127 ± 0.271 | 0.127 ± 0.210 |
| A-HHDM | 0.828 ± 0.166 | 0.665 ± 0.131 | 0.822 ± 0.088 | 0.635 ± 0.101 | 0.550 ± 0.208 | 0.714 ± 0.145 | 0.570 ± 0.172 | 0.759 ± 0.179 | 0.598 ± 0.069 |
| AR-HHDM | **0.912 ± 0.091** | 0.759 ± 0.108 | 0.889 ± 0.062 | 0.731 ± 0.081 | 0.630 ± 0.182 | 0.766 ± 0.128 | 0.651 ± 0.148 | 0.784 ± 0.175 | 0.690 ± 0.064 |
| ARM-HHDM | **0.966 ± 0.030** | 0.833 ± 0.096 | **0.931 ± 0.045** | **0.817 ± 0.061** | **0.708 ± 0.175** | 0.821 ± 0.100 | 0.726 ± 0.118 | 0.835 ± 0.144 | **0.802 ± 0.044** |
| ARMS-HHDM | **0.969 ± 0.027*** | **0.846 ± 0.095** | **0.934 ± 0.044*** | **0.827 ± 0.060*** | **0.724 ± 0.175*** | **0.827 ± 0.095** | **0.762 ± 0.103*** | **0.856 ± 0.138** | **0.823 ± 0.042*** |
| MHDM | 0.780 ± 0.136 | 0.684 ± 0.131 | 0.653 ± 0.081 | 0.535 ± 0.136 | 0.602 ± 0.173 | 0.656 ± 0.116 | 0.627 ± 0.129 | 0.648 ± 0.157 | 0.544 ± 0.118 |
| MHDM (nested evaluation) | **0.962 ± 0.035** | **0.867 ± 0.096*** | **0.901 ± 0.055** | 0.752 ± 0.076 | 0.690 ± 0.174 | **0.846 ± 0.087*** | **0.749 ± 0.108** | **0.916 ± 0.084*** | **0.754 ± 0.050** |
| Friedman statistics (p-value) | 42.63 ($4.10^{-8}$) | 122.47 ($9.10^{-25}$) | 49.25 ($2.10^{-9}$) | 98.83 ($9.10^{-20}$) | 259.25 ($5.10^{-54}$) | 185.54 ($3.10^{-38}$) | 219.3 ($2.10^{-45}$) | 122.49 ($9.10^{-25}$) | 59.86 ($1.10^{-11}$) |
| N° classes | 11 | 25 | 10 | 20 | 54 | 42 | 46 | 29 | 12 |

When computing precision and recall within each formation, the global model achieved performance comparable to the best hierarchical variant across most formations. However, in the two most heterogeneous formations (wetlands and grasslands), the hierarchical design still provided a statistically significant improvement (Friedman $p \leq 10^{-9}$) indicating that its superiority does not stem only from its evaluation advantage.

Figure 4 summarises the precision–recall balance of the five hierarchical variants. Using only the abiotic layer already produced well-performing models, but the addition of RS Products systematically improved both recall and precision, especially for vegetated formations. The inclusion of MSI and SAR embeddings yielded a further gain in both recall and precision.

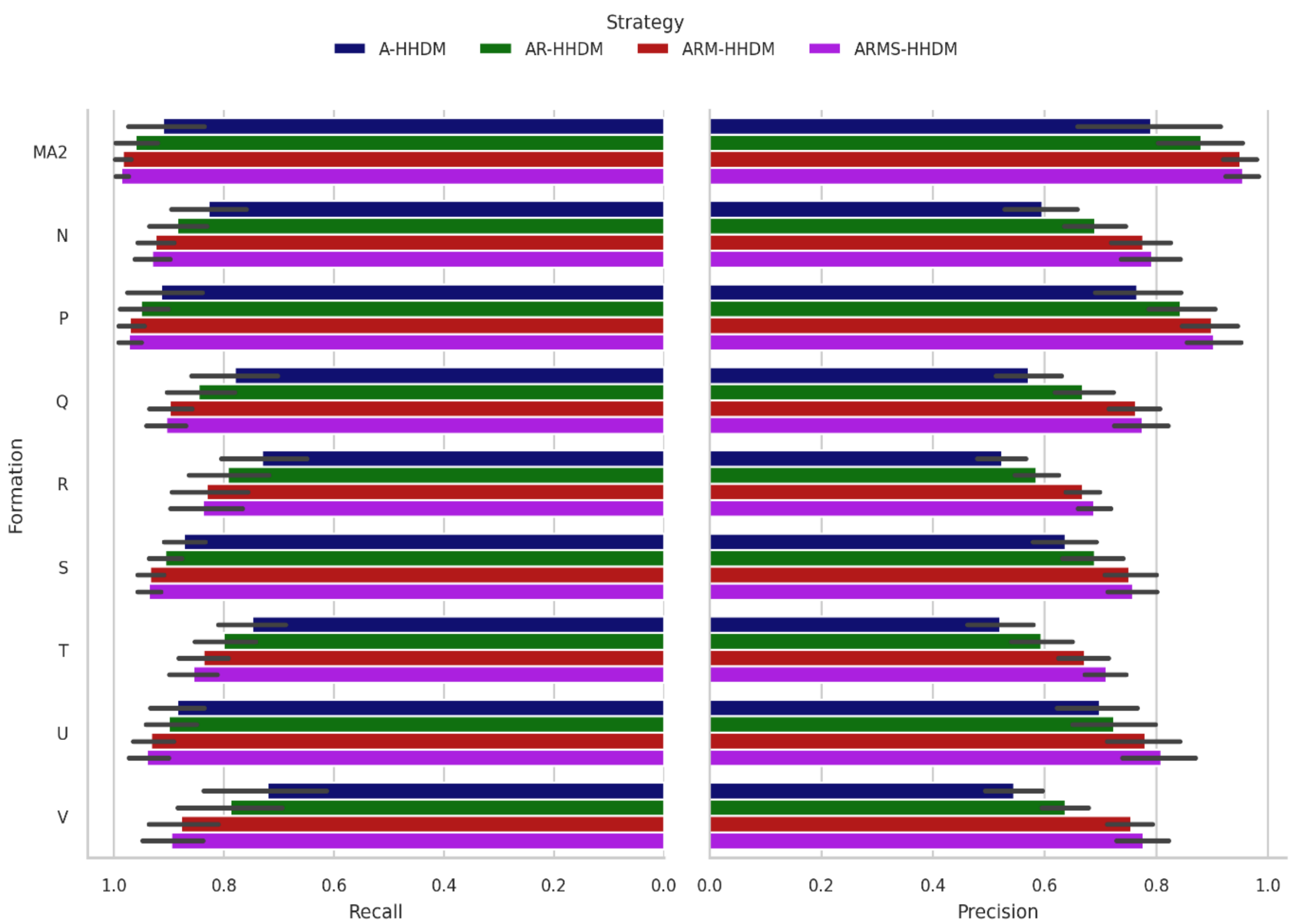

*Figure 5 - Recall-precision trade-offs on the validation datasets across habitat modelling strategies*

## 3.6. Feature attribution

SHAP analyses on the best overall configuration (ARMS-HHDM) showed that abiotic variables explain more than half of the spatial variation in habitat occurrence across all formations. Climatic descriptors, mainly growing-season temperature, annual precipitation and snow cover, were particularly influential in discriminating saltmarsh, grassland and shrubland classes. Topographic proximity to the coast emerged as the most universally important single terrain attribute, though less so inside the salt-marsh formation itself. Geological and fine-scale soil properties contributed modestly but uniformly overall (Figure 6, top).

RS products accounted for up to one half of the explained variance. Hydrological layers, landscape-composition metrics and soil-moisture indices were the leading contributors, followed by vegetation-structure variables. By contrast, the combined contribution of MSI and SAR embeddings remained below ten per cent, indicating that most spectral–textural information captured by the encoders overlaps with that already contained in the RSP variables (Figure 6, bottom).

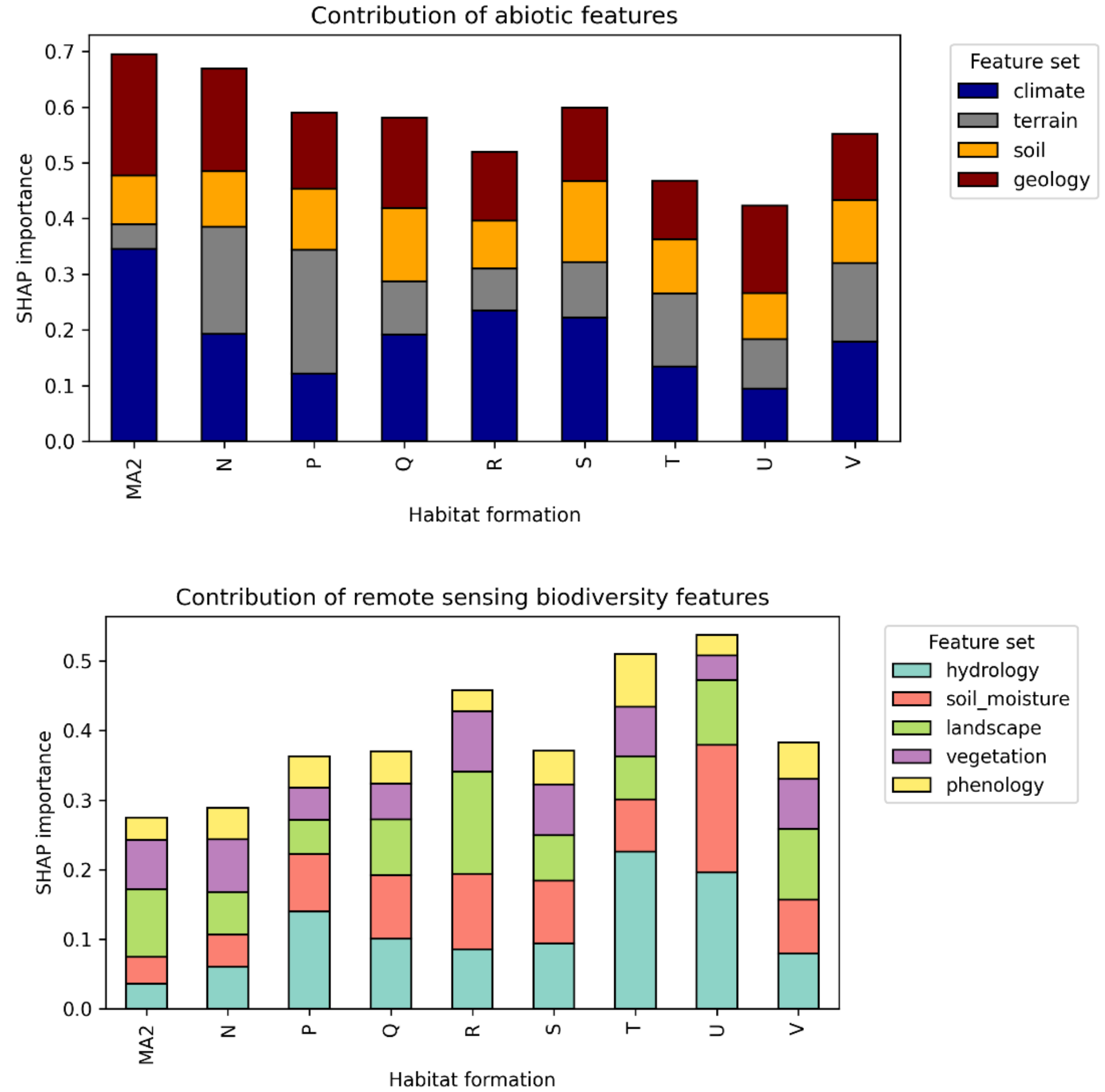

*Figure 6 - Relative contribution of abiotic (top panel) and remote sensing product (bottom panel) sets across habitat formations for the hierarchical habitat distribution modelling (ARMS-HHDM) strategy.*

# 4. Discussion

This study pursued two complementary goals: first, to quantify how strongly different modalities of remote-sensing data improve habitat mapping at fine thematic resolution and fine spatial resolution; and second, to identify the learning configurations (classification schemes, architecture, loss functions and spatial context) that make habitat distribution modelling both accurate at fine spatial resolution and computationally tractable across large spatial extent. By confronting several modelling strategies with more than two hundred habitat classes across the European continent, we draw several conclusions.

## 4.1. Beyond climate: environmental drivers and remote sensing signatures

All learning strategies that relied on environmental predictors outperformed the biogeographic baseline, confirming that statistical models are essential to move beyond the very coarse pictures delivered by reporting data of the habitat directive. Given that biogeographic regions are defined by climate conditions, this result also highlights the importance of landscape-level and local environmental drivers in shaping habitat distributions.

The feature attribution demonstrated that abiotic layers and hydrological properties are the most informative sets of predictors. Indeed, climate, topography and geological history drive large scale habitat range (potential area of occupancy) patterns (Álvarez-Martínez et al., 2018). At fine scale, remotely sensed hydrological conditions including water body extent and inundation frequency as well as soil moisture complements terrain landforms and soil properties in describing micro-climate conditions (Haesen et al., 2023; Kemppinen et al., 2024; Zellweger et al., 2019). Altogether, these variables describe the ecological niches of plant species linked to these habitats (Leitão & Santos, 2019; Thuiller, 2024).

Beyond the local suitability, the emergent structural (e.g., vegetation height, canopy density) and functional (e.g., phenological dynamics) properties of the habitats captured by remote sensing vegetation products further allows the fine-grained discrimination of habitat classes.

Habitats seldom occur as discrete patches but rather within characteristic mosaics shaped by land-use intensity, successional stages and biophysical gradients (Burel, 2003). By representing the relative cover of neighbouring land-cover classes, landscape composition variables help models infer contextual constraints and facilitative relationships that single-pixel predictors overlook. This improves discrimination in ecotonal and heterogeneous landscapes, where transitions and mixed pixels are ecologically meaningful (Kupfer et al., 2006; Opdam et al., 2001) (Turner & Gardner, 2015).

Multispectral and radar embeddings contributed a smaller share of the total variance, yet their incremental value was systematic, particularly in mosaics where natural and anthropogenic land covers form intricate spatial textures. The fact that SAR was most helpful along coasts and in sparsely vegetated outcrops confirms its complementarity to optical imagery in situations dominated by surface roughness or dielectric contrasts (Torres et al., 2012b).

## 4.2. Hierarchy matters more than exclusivity

The way mutual exclusivity was enforced among classes during training proved decisive. Hierarchical models benefited from the focus on specific formations to learn better rules (tree-based approaches) and class boundaries (neural networks) to distinguish habitat classes at level 3. As a result, hierarchical strategies consistently produced the best precision without sacrificing recall. This strategy is particularly useful when habitat classes from different formations share similar abiotic features and spectral signatures.

Many coastal habitats (e.g. cliffs, saltmarshes) share similar properties to other habitats from inland formations. In this particular case, our results show that considering the optical and radar imagery enables the distinction of these habitats from all other formations. This highlights both the importance of the spatial context and the rich information in satellite imagery that is not captured by land cover variables.

Some natural habitat classes are hard to distinguish from man-made habitats. In this case, incorporating land use information is crucial and can be done seamlessly in the proposed hierarchical framework by applying crosswalk rules (Si-Moussi et al., 2024).

## 4.3. Self-supervised satellite encoders prevail

Embeddings produced by self-supervised EO-FMs out-performed the supervised model that had been trained explicitly on broad habitats. This echoes a growing consensus in the Deep Learning and Earth Observation communities that self-supervised learning pretext tasks, by exposing a network to the full diversity of spectral-textural patterns present in raw satellite archives, extract representations that transfer better to downstream ecological tasks (Reichstein et al., 2019; Xiao et al., 2024). Surprisingly, the choice of pretext task itself and the encoder architecture (CNN, ViT) did not have a strong effect on our task as evidenced by the inconsistent ranking of EO-FMs, without fine-tuning.

In this study, we selected the Dynamic-One-For-All (DOFA) model due to its balanced recall and precision. DOFA is based on a vision transformer which can capture spatial relationships between landscape elements. Other EO-FMs provided similar performances including

SSL4EO ViT variants. However, we selected DOFA due to its explicit encoding of wavelength information, support for multi-modal imagery and active maintenance (Y. Wang et al., 2025). Unlike recent models (e.g., AlphaEarth) (Bodnar et al., 2025) that are provided in the form of geospatial embeddings mixing various modalities, DOFA can separately encode different modalities (e.g., MSI and SAR) thus allowing to quantify their relative contributions without the confounding effect of encoder differences.

Given the recent hype and ever-increasing publication of new foundation models, a benchmarking across a large and diverse set of biodiversity related tasks is needed to guide end users in choosing the most appropriate model (Zhu et al., 2024).

## 4.4. Evidence of multi-scale dependencies

Patch size exerted a second order but significant influence: 1-km windows maximized recall in woody or hydrogeomorphic formations where habitat extent usually exceeds several hundred meters, whereas 500-m windows reduced false positives in heterogeneous grasslands and coastal habitats. This scale dependency suggests that future global pipelines should incorporate a nested architecture able to fuse contextual cues extracted at several nested extents (Tiel et al., 2024).

## 4.5. Learning algorithms, imbalance and ensemble design

Ensemble averaging individual algorithm predictions yielded the lowest coverage error, confirming that the diversity of inductive biases between decision trees and neural networks provides effective variance reduction (Sagi & Rokach, 2018). The trade-off between minority and majority classes, however, remained delicate. In tree models, simple class weighting improved macro-averaged scores unless the minority-to-majority ratio exceeded roughly 1:50, beyond which imbalance correction penalizes common classes. Adaptive losses such as focal loss and LDAM offered a better compromise for neural networks, in line with findings from computer-vision benchmarks with long-tailed label distributions (Cui et al., 2019). Extending such margin-based losses to gradient-boosted trees appears a promising short-term avenue (C. Wang et al., 2020).

## 4.6. Generalizability and transfer

Although the present analysis was anchored in the EUNIS hierarchy, the modelling workflow, spatial blocking, hierarchical exclusivity and multimodal stacking, is generic. EUNIS was selected due to its hierarchical structure, ecological soundness, and broad recognition across Europe, including compatibility with national nomenclatures. Regardless of the classification system or geographic focus, the methodology presented here can support robust habitat modelling across scales and typologies.

## 4.7. Limitations and perspectives

Three caveats deserve explicit mention. First, for computational parsimony we refrained from fine-tuning the EO-FMs, accepting a potential loss of class-specific sensitivity; recent work (Scheibenreif et al., 2024) suggests that even a single epoch of fine-tuning can raise recall by 3–4 % in rare classes. Second, our labels correspond to plot centroids, not polygons, which inevitably introduces positional noise. Future expert- or crowd-derived delineations (Fritz et al., 2017; Wu et al., 2024) could enable true segmentation and area-based habitat validation. Third, despite the spatial blocking protocol, observational density is higher in central and northern Europe than in the Mediterranean or boreal regions. As a result, absolute accuracies may be optimistic in under-sampled ecoregions.

Several technological trends are poised to reshape habitat mapping. Foundation models combining text and imagery (Radford et al., 2021) already allow open-vocabulary recognition of land-cover types (Jain et al., 2025); their extension to ecological classes could relax the dependence on fixed nomenclatures and facilitate integration with citizen-science descriptions.

Upcoming missions will greatly enhance habitat monitoring capacity. Dense optical time series from Sentinel-2 NG and the thermal-infrared LSTM mission will refine discrimination of hydrological and phenological regimes. CHIME (Celesti et al., 2022) will add hyperspectral sensitivity to vegetation and soil traits, while NISAR and BIOMASS(Scipal et al., 2010) will contribute complementary radar observations capturing canopy structure and height. Together, these missions will provide an unprecedented basis for mapping habitat composition, structure, and condition across Europe.

## 4.8. Implications for restoration and monitoring

Looking ahead, the integration of artificial intelligence with multimodal Earth observation offers major opportunities for operational biodiversity monitoring. By delivering fine-grained and thematically rich habitat maps, these approaches can directly support the development of Essential Biodiversity Variables (EBVs) for ecosystem structure and distribution (Pereira et al., 2013), which are central to global monitoring initiatives. At the European level, such tools are highly relevant for tracking progress toward the EU Nature Restoration Law and the EU Biodiversity Strategy for 2030, while globally they align with the monitoring needs of the Kunming–Montreal Global Biodiversity Framework. Beyond static mapping, the ability to incorporate hierarchical structure, microenvironmental predictors, and dynamic variables such as phenology provides a foundation for monitoring not only habitat extents but also their heterogeneity and change through time. This opens the door to more realistic assessments of restoration outcomes as well as better anticipation of pressures from forestry, agriculture, and urban expansion. Realising this potential will require sustained efforts to harmonise field data, advance EO foundation models, and ensure interoperability with policy frameworks.

### Data availability

For reproducibility purposes, spatial habitat observation records are available at Zenodo (https://doi.org/10.5281/zenodo.16944381) under a CC-BY 4.0 license. Additional metadata and the full species composition per vegetation plot is only available by request to the EVA database administrators (https://euroveg.org/eva-database/obtaining-data).

### Code availability

All of the code used in this study is openly available at the GitHub repository: https://github.com/bettasimousss/EOHabitatModeling including environment settings, scripts for data preparation, embedding/inference with multiple foundation models, downstream model training and configuration files for reproducibility.

### Author contributions

SSM, WT, and SM conceived the study. SL and SSM prepared the environmental and remote sensing datasets. BJA, FA, JCS, and SH curated the habitat data. SSM led the comparative analyses. SSM wrote the first draft of the manuscript with input from WT, SH and SM. All authors contributed to the interpretation of results and to revising the manuscript.

### Funding

This work was carried out in the frame of the EO4DIVERSITY project funded by the European Space Agency as part of its Biodiversity+ Precursors programme of the ESA-EC Flagship Action on Biodiversity and Vulnerable Ecosystems. We also acknowledge funding from the Horizon Europe OBSGESSION project (No.: 101134954).

### Acknowledgements

The authors thank Joaquim Estopinan and Vincent Miele for insightful discussions on earth observation foundation models; Milan Chytrý, Gianmaria Bonari, Ute Jandt, Behlül Güler, and Jonath Lenoir for their insightful comments on the results; the European Vegetation Archive data providers and curators for providing the habitat data; and the EO4DIVERSITY consortium for their helpful feedback.